\documentclass{article}

\usepackage{arxiv}

\usepackage[utf8]{inputenc} 
\usepackage[T1]{fontenc}    
\usepackage[hyphens]{url}          
\usepackage{hyperref}       
\usepackage{booktabs}       
\usepackage{amsfonts}       
\usepackage{nicefrac}       
\usepackage{microtype}      
\usepackage{lipsum}
\usepackage{graphicx}

\usepackage{tikz}
\usepackage{tikz-qtree}
\usepackage{subcaption}

\newcommand{\q}[1]{``#1''}

\title{Bridging the Communication Gap: Evaluating AI Labeling Practices for Trustworthy AI Development}

\author{
 Raphael Fischer \\
  Lamarr Institute for ML and AI\\
  TU Dortmund University\\
  Dortmund, Germany\\
  \texttt{raphael.fischer@tu-dortmund.de} \\
   \And
 Magdalena Wischnewski \\
  RC Trustworthy Data Science and Security\\
  TU Dortmund University\\
  Dortmund, Germany\\
  \texttt{magdalena.wischnewski@tu-dortmund.de} \\
  \And
 Alexander van der Staay \\
  Enterprise Computing\\
  TU Dortmund University\\
  Dortmund, Germany \\
  \texttt{alexander.staay@tu-dortmund.de} \\
  \And
 Katharina Poitz \\
  Lamarr Institute for ML and AI\\
  TU Dortmund University\\
  Dortmund, Germany \\
  \texttt{katharina.poitz@tu-dortmund.de} \\
  \And
 Christian Janiesch \\
  Enterprise Computing\\
  TU Dortmund University\\
  Dortmund, Germany \\
  \texttt{christian.janiesch@tu-dortmund.de} \\
  \And
 Thomas Liebig \\
  Smart City Science \\
  TU Dortmund University\\
  Dortmund, Germany \\
  \texttt{thomas.liebig@cs.tu-dortmund.de}
}

\begin{document}
\maketitle
\begin{abstract}
  As artificial intelligence (AI) becomes integral to economy and society, communication gaps between developers, users, and stakeholders hinder trust and informed decision-making.
  High-level AI labels, inspired by frameworks like EU energy labels, have been proposed to make the properties of AI models more transparent.
  Without requiring deep technical expertise, they can inform on the trade-off between predictive performance and resource efficiency.
  However, the practical benefits and limitations of AI labeling remain underexplored.
  This study evaluates AI labeling through qualitative interviews along four key research questions.
  Based on thematic analysis and inductive coding, we found a broad range of practitioners to be interested in AI labeling (RQ1).
  They see benefits for alleviating communication gaps and aiding non-expert decision-makers, however limitations, misunderstandings, and suggestions for improvement were also discussed (RQ2).
  Compared to other reporting formats, interviewees positively evaluated the reduced complexity of labels, increasing overall comprehensibility (RQ3).
  Trust was influenced most by usability and the credibility of the responsible labeling authority, with mixed preferences for self-certification versus third-party certification (RQ4).
  Our Insights highlight that AI labels pose a trade-off between simplicity and complexity, which could be resolved by developing customizable and interactive labeling frameworks to address diverse user needs. 
  Transparent labeling of resource efficiency also nudged interviewee priorities towards paying more attention to sustainability aspects during AI development.
  This study validates AI labels as a valuable tool for enhancing trust and communication in AI, offering actionable guidelines for their refinement and standardization.
\end{abstract}


\section{Introduction}\label{sec:intro}

\renewcommand{\arraystretch}{1} 

As artificial intelligence (AI) advances, companies are increasingly integrating it into their daily operations.
This involves different stakeholders, such as software developers, domain experts, and project leaders, who need to reach agreements despite their very different levels of expertise.
To ensure the trustworthy \cite{chatila_trustworthy_2021} and sustainable \cite{rohde_broadening_2024,van_wynsberghe_sustainable_2021} use of AI, it is imperative to bridge the communication gaps between the diverse parties that develop, use, or are affected by AI solutions.
Examples of these gaps include limited technical understanding (even on the developer side \cite{kaur2020}) and unrealistic expectations \cite{piorkowski_how_2021}, which can result in misuse and disuse of the technology \cite{parasuraman1997}. 
Whether stakeholders aim to use AI services \cite{lins2021artificial,elger2020ai} or develop custom machine learning (ML) models, comprehending AI behavior and its practical implications is crucial but nontrivial.

To make informed decisions, stakeholders thus require a comprehensible form of communication about practical AI properties and performance trade-offs--such as resource demands versus predictive quality \cite{fischer_metaqure}).
Established forms of reporting such as papers and result databases mostly address experts and are biased towards focusing on predictive performance \cite{fischer_towards_2024,birhane_values_2022}.
To foster resource-awareness and be transparent towards audiences that are less proficient in AI, Fischer et al. \cite{fischer_unified_2022} proposed more comprehensible, high-level \emph{labels}.
In analogy to established systems such as the EU energy labels, these AI labels aim to inform about the intricate trade-offs occurring among different AI models without presupposing any more profound understanding of ML \cite{morik_yes_2022}. 
While the idea of AI labeling was positively acknowledged \cite{pimenow_challenges_2024,duran_gaissalabel_2024,castano2023exploring,heaton2023explainable} to possibly be an \q{excellent tool} \cite{genovesi2022acknowledging}, an empirical evaluation is missing and several works even questioned the effectiveness of role model systems \cite{sola_promoting_2020,PETERS2024100380} and AI trust seals \cite{wischnewski2024seal,scharowski_certification_2023}.

\begin{figure}
    \hfill
    \begin{minipage}[c]{0.4\linewidth}
        \centering
        \includegraphics[width=0.48\linewidth]{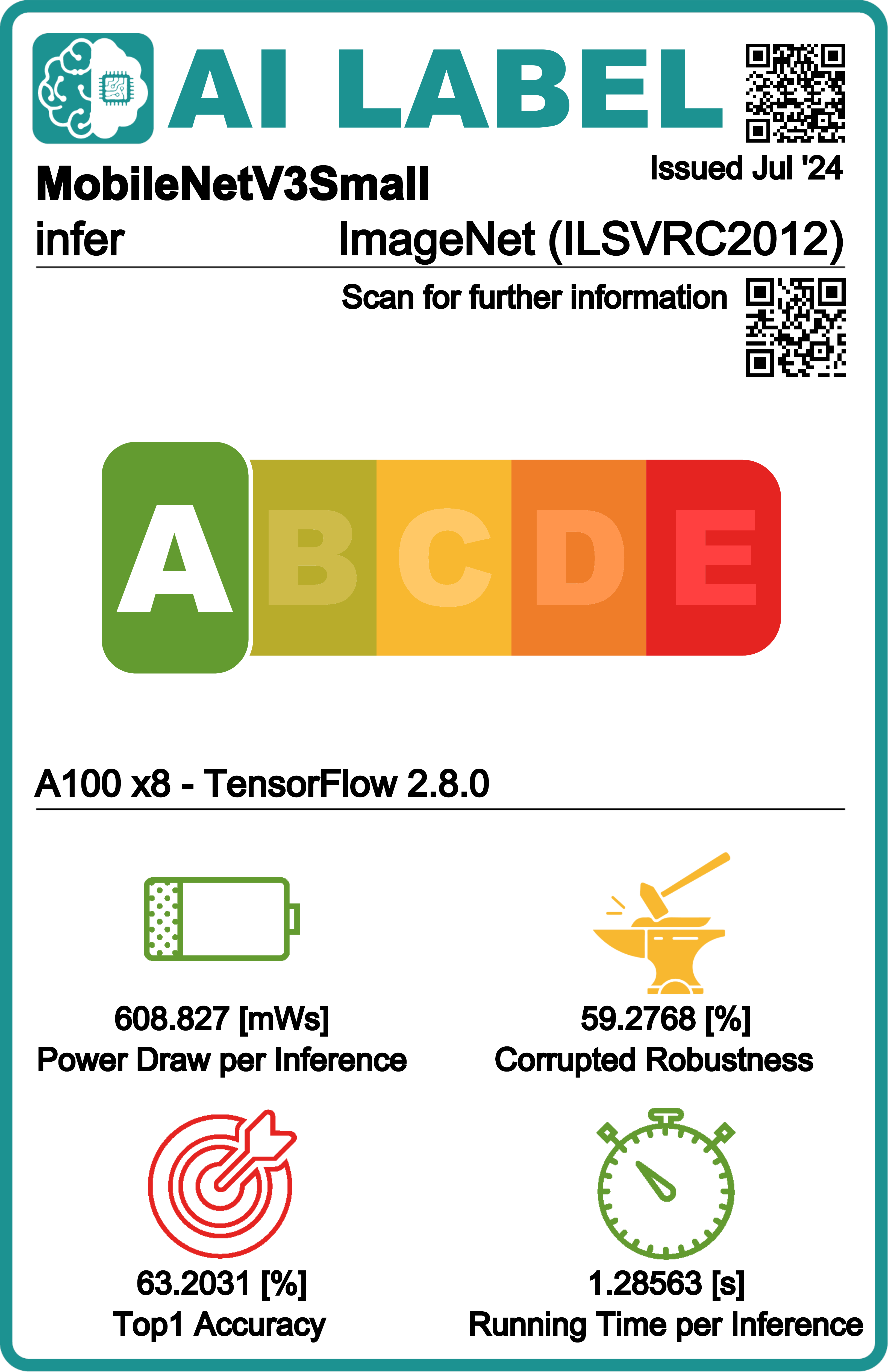}
        \hfill
        \includegraphics[width=0.48\linewidth]{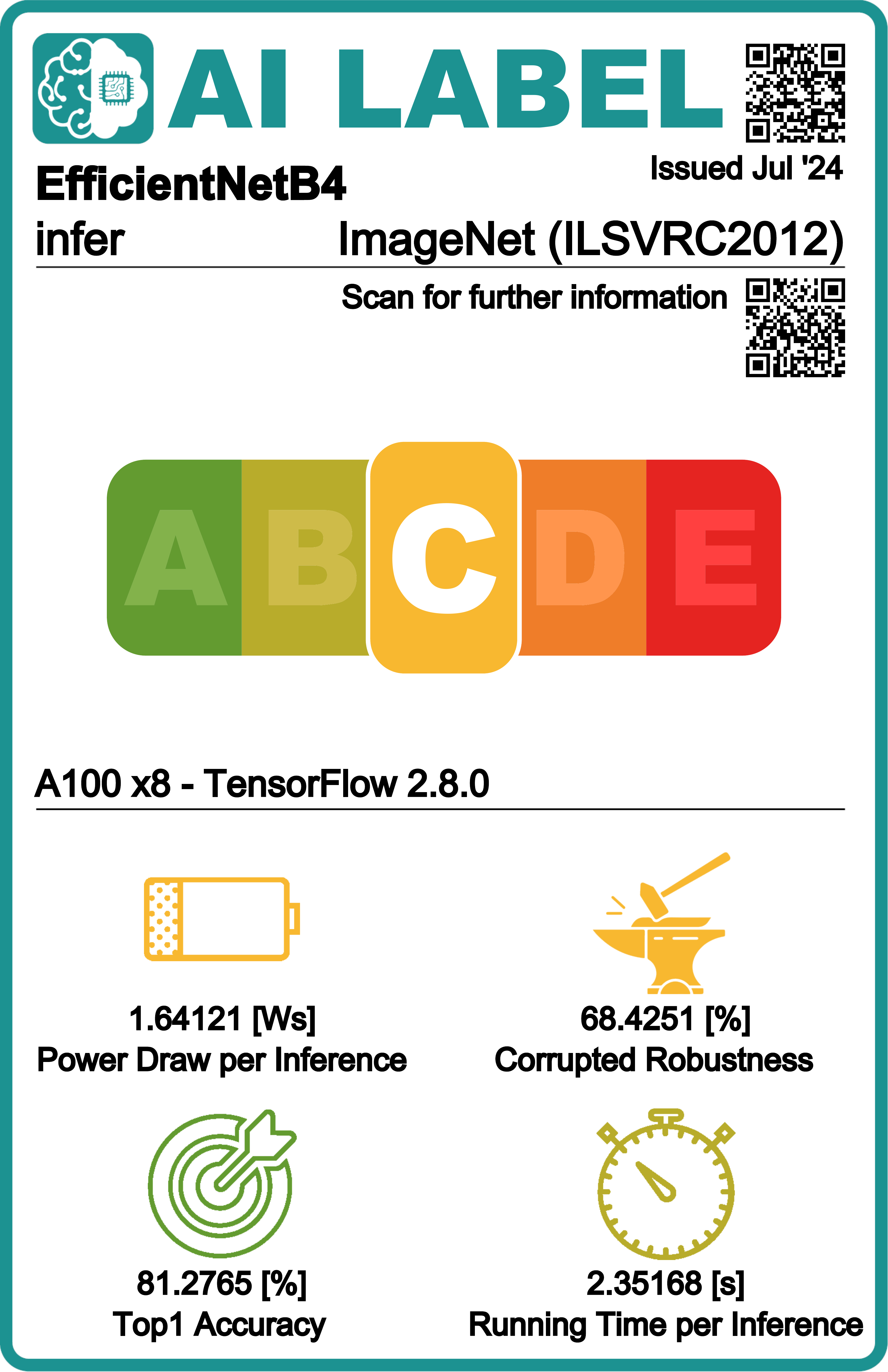}
        \caption{Prototype AI labels, as generated by \texttt{STREP} \cite{fischer_towards_2024} and shown during interviews.}
        \label{fig:labels}
    \end{minipage}%
    \hfill
    \begin{minipage}[c]{0.55\linewidth}
        \centering
        \captionof{table}{First impressions when facing the presented AI labels}
        \begin{tabular}{c|p{0.87\linewidth}}
            ID & Interview quote (in reaction to labels) \\
            \hline
            I3 & \q{looks like I'm looking for a washing machine at the store. Very consumer-friendly, if what it says is true} (p.~84) \\ 
            I5 & \q{for the first entry it is perfect, to see which model I get started with} (p.~228) \\
            I7 & \q{the greatest benefit that I can see right away is directed at the decision-makers and customers} (p.~74) \\
            I9 & \q{the big advantage that your label has is that it helps me to land on a decision immediately. So it reduces a lot of the time expenditure} (p.~219) \\
            I16 & \q{looks like I can judge quite well how energy-intensive the whole thing is, how reliable the whole thing is, how fast the whole thing is} (p.~34) \\ 
        \end{tabular}
        \label{tab:impressions}
    \end{minipage}
    \hfill
\end{figure}

Hence, with this work, we address the need for an in-depth evaluation of AI labeling by conducting an interdisciplinary user study, focusing on the following main research questions:
\begin{enumerate}
    \item[RQ1]: Who is interested in AI labeling and what are their problems with AI technology?
    \item[RQ2]: What are the practical benefits and limitations of AI model labeling?
    \item[RQ3]: How are AI labels perceived compared to other forms of reporting?
    \item[RQ4]: How do AI labels and the corresponding certifying authority affect the trustworthiness of AI systems?
\end{enumerate}
To answer our research questions, we conducted semi-structured interviews with 16 participants from various application domains, covering diverse levels of AI expertise.
Besides discussing their daily work with AI, we also confronted them with AI labels as displayed in Figure \ref{fig:labels} to assess their advantages and limitations and drew comparisons with other forms of reporting -- we highlight some first impressions in Table \ref{tab:impressions}.
Following thematic analysis, we developed an extensive code system around our research questions (over 1000 occurrences) and now contribute an in-depth discussion of the corresponding positions.

Our findings demonstrate a strong interest in AI labeling and practical benefits for connecting ML experts with less-informed users despite occasional misunderstandings and concerns about technical complexity. Participants recognized clear practical benefits of AI labels, including their potential to enhance decision-making, facilitate communication, and promote knowledge transfer between ML experts and non-experts. Moreover, our study suggests that AI labels can act as "nudges," fostering more informed and sustainable decision-making.
Our interviewees highlighted the importance of usability, sustainability, and customizable labeling formats tailored to diverse audiences.
Importantly, our analysis implies that labeling should not be understood as a `one-fits-all' solution. Instead, future efforts should focus on developing interactive frameworks for generating AI reports that cater to specific user needs.
For that, our research questions offer a foundation for refining and standardizing AI labeling procedures to achieve this goal.
Our work validates the theoretical concept of AI labels for practical feasibility, showcasing its capability to bridge communication gaps and even benefit sustainable development.
Moreover, we root our research in Open Science practices, making all supplementary results, such as the transcripts and coding system available at \url{www.github.com/raphischer/labeling-evaluation}.
Based on our findings, we deem AI labeling a central communication format for bridging gaps in the field and making AI systems more trustworthy and sustainable.

\section{Related Work}
\label{sec:rw}


\paragraph{\textbf{On the Current Challenges in AI Development}}
\label{ssec:rw:ai_development}

Until recently, incorporating AI into business required skilled ML engineers and developers who analyzed the business use case and data at hand to train custom models.
Small and medium-sized enterprises often struggled to keep pace in the race to make business and profit with AI, as it required substantial upfront investments in hardware and human expertise, often without guaranteed returns \cite{boag2018dependability}. 
The last years, however, brought forth a paradigm shift that centers on the availability of AI-as-a-service (AIaaS) \cite{lins2021artificial,elger2020ai}.
It enables businesses to access AI capabilities via cloud services and easy-to-use interfaces, which differ in their levels of abstraction and customizability. 
The most well-known examples from this new era are large-scale language models \cite{llm_brown_few_shot_learners} like ChatGPT, whose extreme internal complexity is hidden behind online prompt interfaces.
Apart from the access to pre-trained models, a variety of services exist, like local model deployment, fine-tuning, or infrastructure services~\cite{lins2021artificial}.

Whether AIaaS or on-premises ML for developing custom models -- AI systems can benefit diverse business domains and use cases, which inevitably leads to knowledge and communication gaps.
Classically, these gaps occur between ML engineers and application experts, which, for example, complicates the identification and prioritization of AI use cases in business and economy \cite{fischer_wilo}.
Identifying various communication gaps in AI development, Piorkowski et al. single out knowledge, establishing trust, and setting expectations as prominent gaps.
Among the central takeaways, the authors point to the importance of customized documentation, which needs to be tailored for the target audience \cite{piorkowski_how_2021}.
Another recent meta-summary based on nearly 5000 ML practitioners concluded that education and good software engineering practices must be advanced to ease the incorporation of ML capabilities \cite{nahar_meta-summary_2023}.
In interactive ML, non-experts were also found to struggle with misunderstanding, for example, in terms of predictive model quality \cite{yang2018}.

With AIaaS, companies do not necessarily need ML engineers to gain access to AI; however, the behavior and practical implications of using these services are hard for non-experts to grasp.  
The lack of demonstration of value that AIaaS might bring to users, in addition to the opacity of costs and other potential drawbacks, represent two significant inhibiting factors to the adoption of AIaaS \cite{pandl2021drivers}. 
It is therefore imperative for AIaaS providers to communicate potential inaccuracies of ML models to increase customers' awareness of potential pitfalls \cite{brecker2023artificial}. 
Improving AI literacy might help to bridge existing communication gaps. However, it is still unclear what AI literacy really encompasses (e.g., \cite{ng_conceptualizing_ai_literacy_2021}).
All these works are evidence that a prime difficulty for using AI in business lies in the way AI developers and AIaaS providers document and report on their advances and models -- currently, understanding them requires profound expertise in the field.

\paragraph{\textbf{On Types and Pitfalls of AI Reporting}}
\label{ssec:rw:ai_reporting}

Reporting is vital for bridging the aforementioned communication gaps between ML experts and less informed practitioners \cite{fischer_towards_2024}.
Scientific publications are well established, however often verbose and incomprehensible for non-experts.
Grey literature like online blogs address a broader target audience, however, follow the respective author's subjective focus and bias
Moreover, benchmarks summarize the performance of state-of-the-art models for specific learning tasks (e.g., adversarial training \cite{croce2020robustbench}) and open databases like Papers With Code \cite{paperswithcode} or OpenML \cite{vanschoren2014openml} list a variety of such results overviews.
However, such comparisons are usually biased, overly focusing on predictive model capabilities \cite{birhane_values_2022} for the respective task and neglecting other aspects such as resource efficiency \cite{fischer_unified_2022}.
For easier comprehension and comparison of ML models, model cards \cite{Mitchell/etal/2019a} and fact sheets \cite{arnold2019factsheets} were proposed (and in the latter case, even patented \cite{arnold_generation_2022}).
While offering valuable additions \cite{pushkarna2022,crisan2022}, unfortunately, these model summaries are still rather technical and cannot be automatically generated \cite{fischer_towards_2024}.

For informing a less knowledgeable target audience, the concept of even more abstract AI labels \cite{morik_yes_2022} was introduced.
In analogy with the European Union (EU) energy \cite{eu_energy_label_2024} or Nutri-score labels \cite{sante_publique_france_nutri-score_2024}, they are supposed to convey only the most important practical aspects of AI models and hide away the complexity of ML.
Theoretically, this allows for swiftly learning about model properties and trade-offs, for example, between model quality and resource consumption \cite{fischer_unified_2022}.
Labels were claimed to aid with sustainable and trustworthy AI development \cite{fischer_towards_2024} and several works have praised the concept \cite{pimenow_challenges_2024,heaton2023explainable,genovesi2022acknowledging} or even presented their own adaptions of the idea \cite{duran_gaissalabel_2024,castano2023exploring}.
However, a proper evaluation of labeling benefits is missing -- well-established role model systems are not without limitations \cite{sola_promoting_2020,PETERS2024100380} and the effectiveness of labeling for increasing trust is also at question \cite{wischnewski2024seal,scharowski_certification_2023}.

\paragraph{\textbf{On the Broader Context of Trustworthiness}}
\label{ssec:rw:trust}
From an interdisciplinary viewpoint, trust in a particular technology is a key prerequisite for not only the general uptake of systems but also their appropriate use \cite{lee_trust_2004,madhavan2007,parasuraman1997}.
Accordingly, there is a great endeavor to align the development of AI with human rights and ethical values, or in short, make AI trustworthy \cite{chatila_trustworthy_2021}.
To that end, users' trust should be appropriately calibrated to a system's actual trustworthiness \cite{wischnewski2023}. 
For cloud services like AIaaS, a duality of trust can be observed, meaning that trust in technology should be complemented by organizational trust in the AIaaS provider \cite{lansing2016trust}.

Several other research areas like AI accountability \cite{novelli_accountability_2024,hauer_overview_2023}, responsibility \cite{baum_responsibility_2022,dignum2019responsible}, or explainability \cite{LANGER2021103473} are closely connected to the strive for trustworthiness. 
These discussions also fueled the recently passed AI regulations, such as the EU AI Act \cite{EU_AI_Act_2024} as well as the United States' California initiatives \cite{newsom2023executive} and presidential executive order \cite{biden2023executive}.
As the \q{world’s first rules on AI}~\cite{parliament_step_2023}, they present a risk-based regulation approach and demand AI systems to be designed as safe, transparent, traceable, non-discriminatory, and environmentally friendly.
The last aspect is of additional importance, as AI possesses the power to make our world more sustainable \cite{vinuesa_role_2020}, however also negatively impacts our society and planet \cite{van_wynsberghe_sustainable_2021}, for example due to extreme CO$_2$ emissions which for generative AI can be as high as 1 gram per query \cite{luccioni_watts}.

Lastly, certification and \q{systematic quality assurance}~\cite{schmitz_why_2022} of AI systems, while not generally demanded by regulatories like the AI Act, are central for establishing trust.
AI labels were also claimed to potentially be beneficial for certification \cite{morik_yes_2022} and improving trust \cite{fischer_towards_2024}.
Labeling approaches indeed offer a promising means to foster trust by addressing information asymmetry between two parties (e.g., ML engineers and users), as centrally discussed in signaling theory.
For users, model labels could (a) convey information on properties (i.e., function as “signals which are actions that parties take to reveal their true type” \cite{kirmani2000},~p.~66) and (b) signal transparency intentions of the provider, possibly increasing users' trust. 
Furthermore, the elaboration likelihood model (ELM) \cite{petty1984} from persuasion literature highlights that labels (a) may serve as cues for peripheral processing, allowing users to assess key aspects of an AI system quickly, or (b) could encourage central route processing, prompting deeper consideration of trade-offs between properties \cite{lowry2012}.

Putting these theoretical considerations into practice, however, gives mixed results.
Kim et al. \cite{kim2008} first found labels not to increase trust in electronic commerce; however, years later, they reported high effectiveness of labeling \cite{kim2016}.
Similarly, trust labels for AI were demonstrated to increase users trust \cite{scharowski_certification_2023}, however the study by Wischnewski et al. \cite{wischnewski2024seal} did not find evidence of increased trust.
The authors base their result interpretation on the trust tipping point theory \cite{adam2020}, suggesting that labels are most effective for moderately trustworthy systems, while their impact diminishes for systems at either extreme of the trust spectrum.
In conclusion, while labels hold potential, their trust-enhancing effects are not universal and require careful evaluation.

\section{Methods}\label{sec:methods}
To evaluate the concept of AI labeling, we started out by formulating our central research questions (cf., Section \ref{sec:intro}) and obtaining ethical approval from the ethics committee of the University of Duisburg-Essen Computer Science faculty (ID: 2407SPWM1293).
To actively foster reproducible research, we make all results, including the interview guide, transcripts, code system, and visualization scripts, publicly available at \url{www.github.com/raphischer/labeling-evaluation}.

\paragraph{\textbf{Approach and Recruitment}}
For conducting our study, we opened a public call\footnote{\url{https://lamarr.cs.tu-dortmund.de/ml-label-interviews/}} for participants to take part in semi-structured interviews.
We particularly invited developers of AI systems but also indicated openness to anyone generally interested in the concept of AI labeling, which was abstractly teased with an exemplary figure.
The campaign was spread via mailing lists and social media posts on LinkedIn, X, WhatsApp, and Instagram, resulting in a certain level of snowball sampling.
While our social networks are naturally biased in consisting of fellow researchers and business partners, we were successful in recruiting a total of 16 practitioners from different fields, backgrounds, and levels of AI experience\footnote{five participants might be biased due to having worked with at least one of the authors, however they are new to AI labeling.} -- an overview is given in Table \ref{tab:interviewees}.
Their level of AI skill was determined via self-assessment, based on some orientation help in our application form: 
The \emph{beginner} (1 person) has a general idea of but no practical experience with AI, \emph{users} (4) have practical experience with AIaaS, \emph{engineers} (8) have performed basic ML on custom data, and \emph{experts} (3) have extensively trained, deployed, and used ML models\footnote{We also offered a \emph{novice} option for candidates without any AI understanding, however did not receive a respective application.}.
Two of our interviewees hold a doctorate as the highest professional qualification, eight have completed a full master's (or diploma) degree, three have graduated as bachelors (or are certified specialists), and the rest have successfully graduated from high school\footnote{This self-assessment is based on the German qualifications framework.}.

\begin{table}
    \centering
    \caption{Overview of Interview Participants and Their Jobs and Skills}
    \begin{tabular}{lllllll}
    ID & Job Title & Company Type & Employees & Gender & Age & AI Skills \\
    \toprule
    I1 & AI Manager & Industrial Manufacturer & 5000-10000 & male & --- & Engineer \\
    I2 & Researcher & Research Service Provider & 51-200 & male & --- & Engineer \\
    I3 & Software Developer \& Student & IT Service Provider & 5000-10000 & male & 20 & Engineer \\
    I4 & Student & University & 40000-45000 & male & 21 & Beginner \\
    I5 & Software Developer \& Student & IT Services (self-employed) & 1 & male & 22 & Engineer \\
    I6 & Solution Engineer & IT Service Provider & 50-200 & male & 28 & User \\
    I7 & Startup CEO \& AI Developer & AI Service Provider & 2-10 & male & 29 & Expert \\
    I8 & Analytics Platform Manager & Public Service Provider & 1000-5000 & male & 30 & Engineer \\
    I9 & Software Developer & Lottery Service Provider & 50-150 & male & 31 & Expert \\
    I10 & Software Developer & IT Services (self-employed) & 1 & male & 31 & Expert \\
    I11 & Data Scientist & IT \& AI Service Provider & 11-50 & female & 32 & Engineer \\
    I12 & Researcher  & Telecommunications & 201-500 & female & 32 & User \\
    I13 & Development Engineer & Industrial Manufacturer & 5000-10000 & male & 43 & Engineer \\
    I14 & Maintenance Manager  & Public Service Provider & 5000-10000 & male & 46 & User \\
    I15 & Principal Cloud Engineer & IT Service Provider & 51-200 & male & 47 & User \\
    I16 & Software Architect & IT Services (self-employed) & 1 & male & 48 & Engineer \\
\end{tabular}
    \label{tab:interviewees}
\end{table}

Every registered application resulted in an interview, for which the participants were compensated with 15€.
Written consent was obtained before the meeting, explaining how participants' identity will be protected by anonymization.
The interviews were conducted via Zoom and only the audio data was saved and analyzed.
All participants received the interview materials after the completed interview and are frequently updated on our study progress. 

\paragraph{\textbf{Interview Structure}}
Guided by our four research questions, we developed an internal interview guide that consisted of four parts.
At the beginning of each interview (part one), we asked participants to introduce themselves, explain how their work relates to AI, and describe difficulties they face in their daily business.
In a second step, the interviewees were initially presented with a prototype AI label, which was generated with the \texttt{STREP} software \cite{fischer_towards_2024}.
It features properties of MobileNetV3Small \cite{Howard_2019_ICCV}, a popular image classification model that is usually used in pretrained form, either locally or as-a-service.
After discussing first impressions and explaining some concepts (if required), we subsequently showed participants a second label featuring an EfficientNet model \cite{pmlr-v97-tan19a}.
Given both labels, as displayed in Figure \ref{fig:labels}, participants were asked to compare the information on both labels, explain which aspects they found helpful and confusing, and comment on what they would change about the label. 
In the third part, we investigated how interviewees reacted to different types of reporting (see Section \ref{ssec:rw:ai_reporting}).
For that, we presented them different reports about MobileNetV3, namely the associated research paper \cite{Howard_2019_ICCV}, the model card on Hugging Face\footnote{\url{https://huggingface.co/qualcomm/MobileNet-v3-Small}}, a blog article \footnote{\url{https://towardsdatascience.com/everything-you-need-to-know-about-mobilenetv3-and-its-comparison-with-previous-versions-a5d5e5a6eeaa}}, the documentation of Keras \footnote{\url{https://keras.io/api/applications/mobilenet/}}, the Papers With Code results\footnote{\url{https://paperswithcode.com/method/mobilenetv3}} \cite{paperswithcode}, and an exemplary fact sheet of IBM\footnote{\url{https://aifs360.res.ibm.com/examples/max_object_detector}} (which unfortunately are only available for IBM products).
Interviewees were then asked to compare the approaches and describe how they would use them in their daily work -- an overview can be found in the Appendix (Figure \ref{fig:reporting_overview}).
Lastly, we opened a conversation around trustworthiness in the context of AI labeling.
The more specific questions in that part discussed possible providers or authorities for labeling models (i.e., who to trust with such a process) as well as the interviewees' position towards certification and regulation.
For each interview, at least two interviewers from the author pool were present\footnote{With our multidisciplinary team, we always paired a computer scientist with either a psychologist or a social scientist.}.

\paragraph{\textbf{Transcription and Coding}}
In a first step, the audio recordings were transcribed via the \texttt{whisper-large-v3} speech recognition model \cite{pmlr-v202-radford23a}, which we deployed locally with the help of the \texttt{Shoutout} tool\footnote{\url{www.github.com/RWTH-TIME/shoutout}}.
Afterward, each interview was manually revised to correct major errors in the transcription.
For our analysis, we followed an inductive coding approach using \texttt{MAXQDA 24.5}.
We discussed individually coded interviews in iterative cycles and mutually refined the coding system.
It was organized in a hierarchical structure, with several overarching code families that individually address aspects of our central questions.
The final system encompasses 136 codes assigned to a total of 1130 text passages, for which an overview is given in Table \ref{tab:codesystem_overview}.
It summarizes the number of codes and occurrences for each of the top-level families as well as their relation to RQ1 -- RQ4 and some exemplary quotes.
The full complexity of our code system is displayed in the Appendix' Figure \ref{fig:full_code_system}.
At the final stage of our code system, we had all authors do a re-coding of two interviews reaching about 90$\%$ intercoder agreeability (obtained by the \texttt{MAXQDA} analysis with code frequency agreement).

\begin{table}
    \centering
    \caption{Overview of the Derived Code System with Number of Occurrences and Exemplary Quotes}
    \begin{tabular}{lllll}
    Code Family & RQ & Size & Occ & Quote \\
    \toprule
    General Codes & Q1 & 8 & 63 & \small \q{To use AI [\dots] to counter the shortage of skilled workers} (I14,~p.~4) \\
    Types of Daily Work & Q1 & 9 & 64 & \small \q{develop an app to detect tolerable products in the supermarket} (I10,~p.~4) \\
    AI Use Cases & Q1 & 10 & 41 & \small \q{monitoring the machine condition such that we can make predictions} (I14,~p.~4) \\
    ML Methods & Q1 & 7 & 64 & \small \q{the AI evaluates whether the typed text contains specific data} (I3,~p.~44) \\
    ML Tools \& Brands & Q1 & 8 & 36 & \small \q{I used scikit-learn models and also worked with TensorFlow} (I13,~p.~4) \\
    Requirements on AI & Q1 & 13 & 118 & \small \q{My boss doesn't care much about the process, he wants results} (I13,~p.~160) \\
    Benefits & Q2 & 12 & 140 & \small \q{Your label helps me to decide immediately, it saves a lot of time} (I9,~p.~219) \\
    Limitations & Q2 & 21 & 205 & \small \q{I don't get how the value is included in the overall scoring} (I16,~p.~58) \\
    Property Importance & Q2 & 5 & 64 & \small \q{the primary objectives: reducing time and enhancing accuracy} (I7,~p.~98) \\
    Associations & Q2 & 3 & 31 & \small \q{like I'm looking for a washing machine at the DIY store} (I3,~p.~84) \\
    Target Audience & Q2 & 1 & 10 & \small \q{the addressees are likely to be people who are intensively involved} (I14,~p.~64) \\
    Workflows and Use & Q3 & 12 & 61 & \small \q{different agendas and newsletters as a regular source of information} (I16,~p.~70) \\
    General Comparison & Q3 & 4 & 46 & \small \q{It is time-consuming -- that is the disadvantage of other approaches} (I9,~p.~219) \\
    Who Needs Trust & Q4 & 3 & 26 & \small \q{it helps to understand how the model works if you are a developer} (I13,~p.~20) \\
    Reasons for Trust & Q4 & 9 & 91 & \small \q{if it has a university stamp on it, it seems more trustworthy} (I11,~p.~144) \\
    Dimensions of Trust & Q4 & 11 & 66 & \small \q{trust in AI, or trust in a label -- these are two different things} (I11,~p.~152) \\
    \midrule
    Total &   & 136 & 1130 &  
\end{tabular}
    \label{tab:codesystem_overview}
\end{table}

\section{Results}\label{sec:results}


\subsection{Who Is Interested in AI Labeling and What Are Their Problems With AI Technology? (RQ1)}\label{ssec:results_who_and_why}

\begin{figure}
    \centering
    \includegraphics[width=0.9\linewidth]{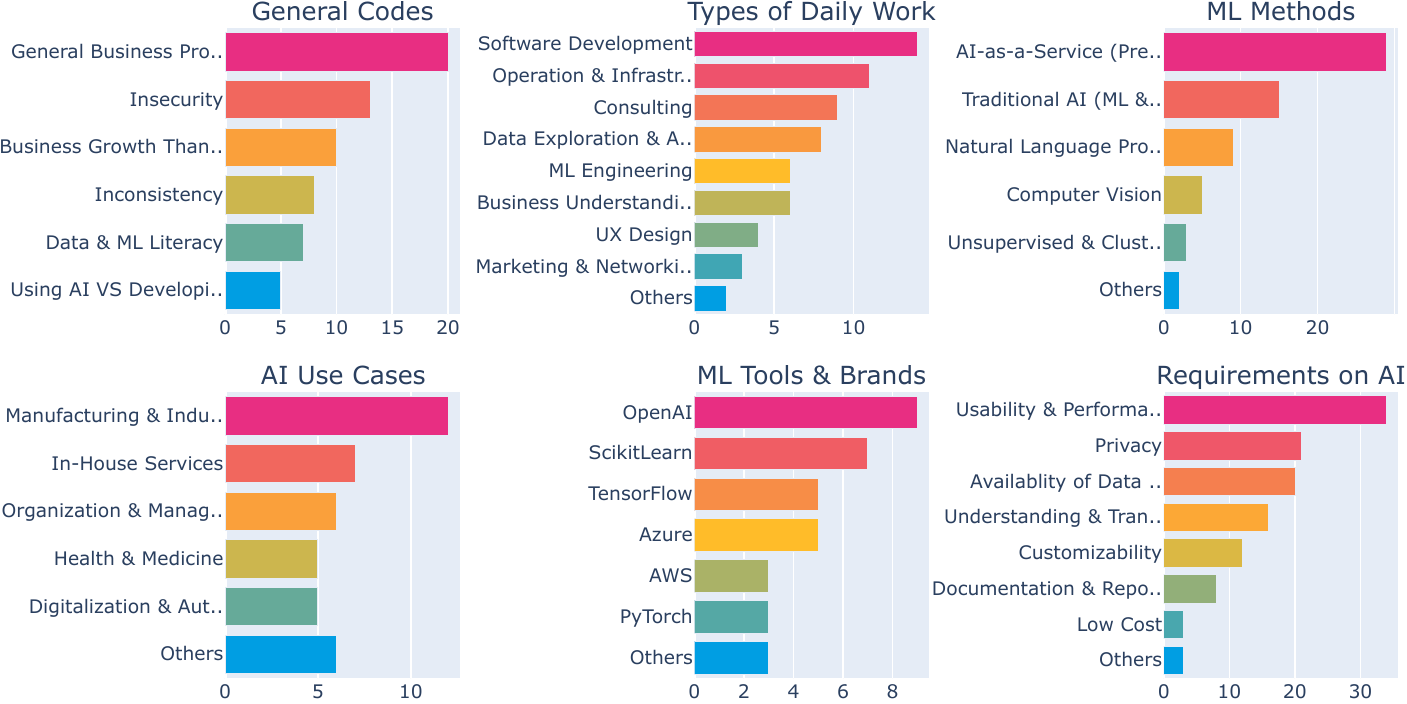}
    \caption{Overview of interviewees' daily work affairs and how they relate to AI (x-axes represents the number of code occurrences).}
    \label{fig:q1_bars}
\end{figure}

To assess possible user groups of AI labels and their needs, we asked participants to describe their daily work in relation to the use of AI methods.
We structured all answers, identifying (1) general problems and positions, (2) participants' types of daily work, (3) ML methods used, (4) AI applications and use cases, (5) specific tools and brands in use, as well as (6) requirements of AI - an overview is given in Figure \ref{fig:q1_bars}.

Firstly, the interviewees talked about several general problems related to the use of AI methods.
I1 described, for example, that it is difficult to \q{get employees on board so that they can actually use the new tools} (p.~26), and I11 mentioned issues with \q{customer communication and expectation management} (p.~50) -- exemplifying communication gaps (seven mentions) as a general business problem (23 in total).
While there seemed a general agreement that AI contributes to business growth (12 mentions), we also encountered various cases of insecurity about the use of AI and inconsistencies in answers, such as, for example, \q{I have quite a few concerns, but on the other hand, I find AI very convenient} (I15,~p.~26). 
Moreover, interviewees also distinguished \q{between AI tools that are used during work or AI tools that are incorporated into products} (I12,~p.~28) -- exemplifying the use scale of a possible labeling approach by also stressing the breadth that would need to be covered.

On similar lines, we encountered a broad spectrum of daily work (see Figure \ref{fig:q1_bars} upper middle) -- software development, infrastructure and operationalization, consulting, as well as data exploration and analysis being most frequently mentioned.
Relating to the use of AI within these different streams of work, we noticed that AIaaS was more frequently encountered than traditionally training models on custom data (upper right), possibly indicating a shift from a predominantly developer- to a customer-based perspective.
We find further support for this speculation statements from our interviewees, such as I7 who stated that \q{when people talk about AI today, they no longer mean deep learning, they mean solely and exclusively large language models} (p.~4).
The many mentions of in-house service applications is likely linked to the this phenomenon (lower left), however manufacturing and industry seem to offer even more use cases which were explicitly mentioned in four of our interviews.
I7 also mentioned that \q{many companies are not yet ready to implement their own deep learning projects} (p.~8), which explains why tools and brands (lower middle) like \texttt{scikit-learn} (for traditional ML, outside of deep learning) and \texttt{OpenAI} (for AIaaS) are most frequently mentioned.

In line with the diversity of possible applications, participants also named a wide range of important requirements for AI systems (lower right), with \q{whether the result works or whether the AI itself is bad} (I5,~p.~98), or in other words, usability and performance, being especially noteworthy.
In this context, most comments (15) referred to performance in terms of predictive quality, while consistency and robustness (4) as well as fast response time (3) received less attention (this can be seen in the full code system of Figure \ref{fig:full_code_system}).
Privacy also played a major role, with many interviewees discussing the importance of data protection and I11 noting that for this reason her company does not develop AI services that have to deal \q{with personal data} (p.~14).
Moreover, data availability played a central role for developers of AI solutions -- \q{the biggest step in ML and training is data collection, and then, feature engineering and data processing} (I9,~p.~40).
Additionally, relating to communication about specific models instead of the actual model performance, we also found understanding and transparency (17), customizability (13), as well as documentation and reporting (8) frequently mentioned as requirements.
This includes the importance of \q{carrying out educational work} for employees (I1,~p.~30), archiving \q{an understanding in terms of what is happening, so that we don't get a black box} (I8,~p.~8), and even \q{finding out what is the right model for my area of application} (I4,~p.~40).

Overall, we conclude that just as user groups are highly diverse, so are their individual challenges with working with AI.
Our practitioners come from diverse backgrounds, and while some train their own models, a larger amount today is simply using AIaaS. 
For the context of AI labeling, this indicates that labels will have to address a diverse group of people with individual challenges.

\subsection{What Are the Practical Benefits and Limitations of AI Model Labeling? (RQ2)}\label{ssec:results_how_to_label}

\begin{figure}
	\centering
	\hfill
	\begin{minipage}[b]{0.54\textwidth}
		\centering
		\includegraphics[width=\textwidth]{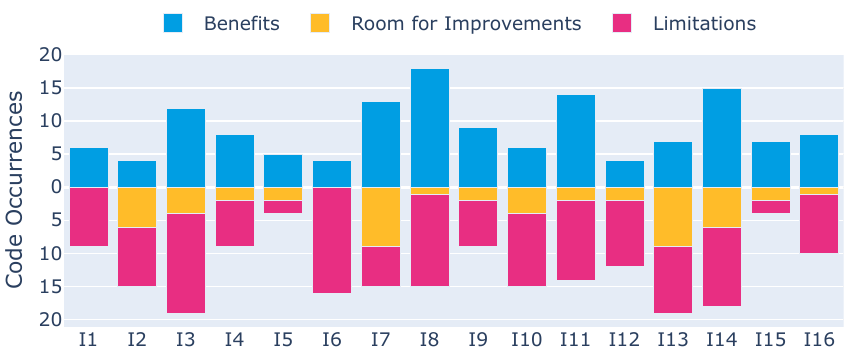}
		\caption{Interviewee sentiment towards labels, based on the number of positive and critical comments.}
		\label{fig:q2_sentiment}
	\end{minipage}
	\hfill
	\begin{minipage}[b]{0.36\textwidth}
		\centering
		\includegraphics[width=\textwidth]{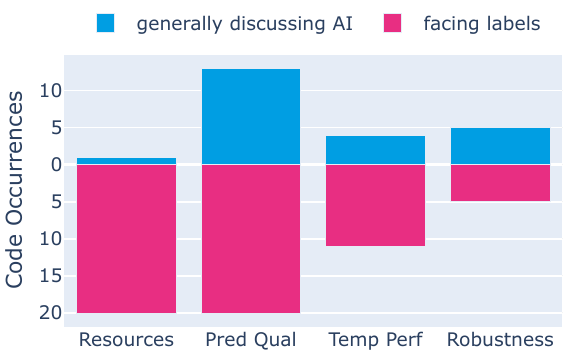}
		\caption{Performance requirements when generally discussing AI and when facing our labels.}
		\label{fig:q2_prop_importances}
	\end{minipage}
	\hfill
\end{figure}

As a possible solution to some of the discussed challenges, we next presented our interviewees with prototypical AI labels. 
A general overview of the interviewee's sentiment towards labels is displayed in Figure \ref{fig:q2_sentiment}, indicating the number of comments on benefits, improvements, and limitations per interviewee. 
Note that the codes on improvements and limitations should not be considered as general positions \emph{against} labeling -- quite the opposite, they can be seen as evidence that the idea is interesting, yet further work and refinement is needed.

\paragraph{\textbf{A Trade-off Question -- Simplicity Over Complexity}}
We observed the general tendency that, on the one hand, interviewees deemed the simplicity of the labels helpful and necessary for decision-making and knowledge transfer, while, on the other hand, they also missed more detailed information. For example, interviewee I1 stated that labels would \q{help in any case -- as there are more and more models, it is increasingly difficult to keep an overview, and the more compact the information is, the better} (I1,~p.~100). 
In this context, I14 perceived our labels as \q{informative at a glance} (p.~56).
In total, we encountered over 50 remarks that described benefits for decision processes, in which labels might be helpful for \q{comparing different models with each other and seeing how well they perform} (I2,~p.~70). 
Interviewees also mentioned advantages in the context of communication and knowledge transfer (20 mentions), with I7 seeing the \q{greatest added value for customer presentations} (p.~174) and I11 stating that customers \q{really like pictures and colors [\dots], so you can always pick people up with categories like red, yellow and green} (p.~106).
On similar lines, one interviewee stated that users \q{like to have the basic information presented immediately. And the AI label, with the color-coded scale, presents an excellent way of doing this} (I14,~p.~170).
Additionally, interviewees saw benefits for transparency (7), advertisement (4), and validation or certification of results (4).

Despite these clear indications of possible benefits, comments also indicated that the abstraction of more complex information, while improving accessibility and fast understanding, could result in confusion or hinder a deeper understanding.
This problem was explicitly mentioned in 19 cases, with I11 anticipating complex factors around models \q{which I don't think you can necessarily cover in a label} (p.~164).
Interviewees directly related this to the target audience: People that have no understanding of ML \q{likely do not really understand} the rather technical presentation of information, which is, instead, only \q{interesting for developers} (I3,~p.~216). 
Overall, our interviewees did not really reach a consensus on who our shown labels address: benefits were seen for \q{people who want to use AI} (I5,~p.~228), \q{decision-makers and customers} (I7,~p.~74), or \q{people who are intensively involved} (I14,~p.~64).

\paragraph{\textbf{Adaptability}}
Possibly bridging the tensions of simplicity and in-depth information, interviewees appreciated the possibility of adapting the label's weights to reflect the user's priorities \cite{fischer_towards_2024}: \q{with a label like this, you have very good opportunities to demonstrate what is important to the developer} (I16,~p.~50).
Interactivity in the labeling procedure was seen as a helpful means to align development with management expectations: \q{if I knew what my boss wanted, I would go to the website, set the weighting, press enter and then I would pull out your label} (I13,~p.~108).
Taking this a step further, our participants also suggested possible improvements to ease the trade-off question, for example, using a fully interactive dashboard \q{that exactly displays those things that are relevant to you} (I7,~p.~166).
I13 also emphasized the need for guiding users with their specific use cases and answer questions like \q{Which model should I use? How do I find my way around?} (p.~72).
On similar terms, I2 (as well as I5 and I8) suggested to better inform on \q{the combination of model quality and application area} (p.~118).

\paragraph{\textbf{Design Benefits, Limitations, and Improvements}}
Going beyond the trade-off question and addressing the label design more directly, interviewees described the labels as generally \q{very consumer-friendly} (I3,~p.~84), and \q{optically appealing} (I7,~p.~94). However, participants also voiced various cases of misunderstanding and confusion, mostly in relation to the displayed performance criteria (lower part of labels), robustness (18), and accuracy (15), specifically.
Here, interviewees could \q{not really imagine what it means} (I4,~p.48) or even mistook robustness (which here relates to adversarial input perturbation \cite{croce2020robustbench}) for a metric describing \q{that the model hallucinates very little} (I6,~p.~86).
I7 and I12 explicitly questioned whether metrics can actually capture the \q{real experience} of using the model (I12,~p.~178) and highlighted the importance of a \q{qualitative, subjective} (I7,~p.~66) evaluation (which unfortunately is non-trivial to embed into ML and AI development \cite{naidu_review_2023}).
Apart from the individual metrics, confusion often originated (20) from the compound scoring aggregation, for example I9 questioned \q{what weights does this have, for example accuracy or power draw, to decide which categories it belongs to} (p.~134). As a solution to these misunderstandings, participants suggested customizing the labeling procedure (20) by adding a second page \q{where the evaluation metrics or the parameters are clearly explained} (I2,~p.~122). 

Despite positive comments about the color-coding (see previous paragraph), we also encountered some general confusion (14) around the relative scoring and colors, such as with I15: \q{what does it mean for me when it is green?} (p.~48). Moreover, one interviewee mentioned that color-scoring is inaccessible for color-blind people, suggesting different letter sizes instead. 
Interestingly, in several cases, the color coding of icons was only registered when the second label was visualized (I7,~p.~78: \q{I didn't realize before that the icons at the bottom were color-coded}). 
Beyond these points, there was also some misconception (12) regarding the model (\texttt{MobileNetV3Small}) and evaluation data (\texttt{ImageNet}), which are well-known in visual computing, however, not so much outside of the domain. Other possibilities of improvement concern the aforementioned color-coding (12), financial cost when deploying the model (7), and additional information on the up-to-dateness of model results (7).

\paragraph{\textbf{Labels as Nudges}}
Facing the labels, we also asked our participants to rate the importance of the displayed properties, namely energy (i.e., resource) consumption, predictive quality (accuracy), temporal performance, and robustness.
Ideally, one would want to have a \q{good mix of all points} (I6,~p.~226), however with 20 mentions each, the first two appear to be most relevant.
The participants here often discussed a trade-off -- on the one hand, \q{power costs money when [the system] is operationalized} (I13,~p.~72), however \q{when accuracy is key, I would also have to accept higher power draw} (I15,~p.~52).
Figure \ref{fig:q2_prop_importances} connects these results with the aforementioned AI requirement of usability and performance, which was analyzed during the more general discussion of daily work (cf. Figure \ref{fig:q1_bars}).
Interestingly, when comparing the respective subcode occurrences with the discussion of property importance on labels, our interviewees appear more resource-aware.
They now discuss model performance trade-offs, with I1 reporting sustainability to be \q{the central driver of [their] corporate strategy} (p.~54) and I2 advocating a general \q{frugality of system complexity} (p.~74).
Echoing the earlier conclusion that labels can aid decision-making processes, we can now add that it is of central importance \textit{which} features a label displays.
If decisions are based on all available information at a glance, then including information such as energy efficiency will inadvertently draw users' attention to these features. 

To summarize, the main benefits of labeling lie in the help with decision processes and knowledge transfer.
The shown labels were generally well-received but can be further improved -- for non-experts, they are still too technical and confusing, while for experts, they do not provide the necessary level of depth.
As a takeaway, there is now evidence that customizability in labeling is key in order to benefit the various target audiences. 

\subsection{How Are AI Labels Perceived in Comparison With Other Forms of Reporting? (RQ3)}\label{ssec:results_competitors}

\begin{figure}
    \centering
    \hfill
    \begin{minipage}[b]{0.45\textwidth}
        \centering
        \includegraphics[width=\textwidth]{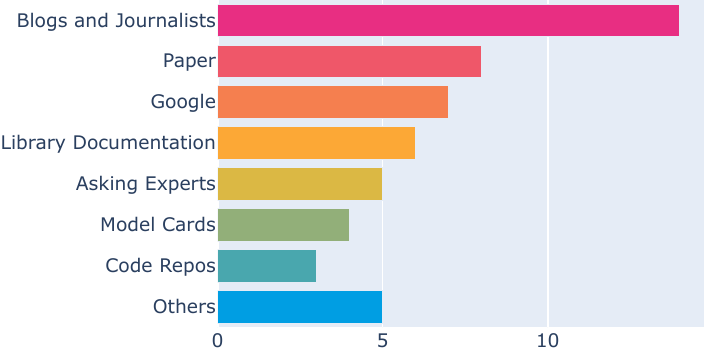}
        \caption{Forms of reporting that are used by interviewees.}
        \label{fig:q3_competitors}
    \end{minipage}
    \hfill
    \begin{minipage}[b]{0.45\textwidth}
        \centering
        \includegraphics[width=\textwidth]{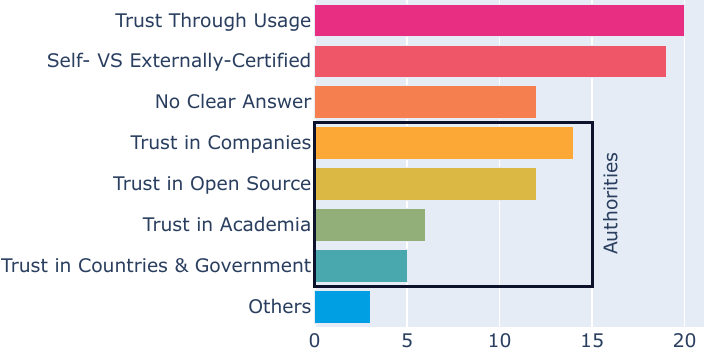}
        \caption{Reasons for trust in the context of labeling and AI.}
        \label{fig:q4_reasons_for_trust}
    \end{minipage}
    \hfill
\end{figure}

We give an overview of actively used report forms in Figure \ref{fig:q3_competitors} and generally found these findings to support our results from RQ2: many interviewees use high-level media and journalistic text the most, however the depth of academic papers is also important -- this clearly reflects the previously mentioned simplicity versus complexity trade-off. 
Supporting low-key access to information, one interviewee mentioned that blogs like Medium are very popular \q{because it's often a practical example that is well explained and easy to work with} (I8,~p.~116). 
I13 and I14 highlighted that educational videos help a lot with getting started with ML and I12 explained that blogs and articles help with learning about \q{trends} and seeing \q{what others do} (p.~186).
Conversely, participants also mentioned issues of trust and reliability for this report type:
\q{What bothers me about Medium is that [\dots] anyone can write anything} (I6,~p.~182).

When comparing other reporting types with AI labels, we found that the access to fast information was especially pronounced.
For example, I9 stated that reading the other formats \q{is time-consuming. [...] That is the disadvantage of all other approaches} (~p.~219) (see also results for RQ2).
This is rooted in the fact that the competitors have \q{significantly more text, significantly more data} (I3, p. 168), which must be consumed and understood: \q{there is quite a lot of complexity involved and I would first have to have a pretty good understanding of it} (I8,~p.~104). 
In contrast, the interviewees appreciated that once you are familiar with the label, \q{you don't even have to read [the sources] anymore -- you just know how good [the models] are.} (I7,~p.~110).
Hence, we conclude that having accessible information appears to be of central importance for daily business and I3 highlighted that \q{none of these [forms of reporting] are as easy to understand as the label} (p.~168).


Contrasting this need for information-at-a-glance, many interviewees also required access to more in-depth information.
I3 stated, for example, that you \q{have to look at [the paper], in any case} and I7 specifically scans them for comparisons \q{with benchmarks and other models} (p.~114).
It was also mentioned that the other forms of reporting are \q{particularly relevant when depth is needed} (I16,~p.~86), or as I3 phrased it: \q{When I read a paper about an AI, I can probably understand it much better than if I just look at the label} (p.~172).
I2 remarked that \q{we need them all -- the difference is, which target groups do I face?} (p.~110) and continues to give examples like papers for scientists, fact sheets and model cards for developers, or blogs for users.
This is in line with the argumentation of Fischer et al., who understand labeling as an addition to the other \q{highly important} forms of reporting \cite{fischer_towards_2024} (p.~6).
Another interviewee suggested seeing labels as some kind of \q{intermediate solution} (I12,~p.~186) for connecting to different groups of people: On the one hand, those who create \q{technical solutions}, on the other hand, \q{product people}, who make them marketable.

In conclusion, the advantages and limitations of different reporting types depend on whether detailed information or just a brief overview is required -- labels in this context are a useful addition as they allow for very fast information intake and can link the other report forms.

\subsection{How Do AI Labels and the Corresponding Certifying Authority Affect the Trustworthiness of AI Systems? (RQ4)}\label{ssec:results_trustworthiness}

Establishing trust in AI systems plays a central role when considering the communicative process of labeling and reporting.
We generally observed two perspectives when discussing trust in the context of AI labels.
On the one hand, interviewees mentioned trust in the labels themselves, as well as in possible issuing authorities.
On the other hand, interviewees mentioned the suitability of labels for establishing trust in AI systems, aptly summarized by I11, who differentiated between \q{trust in AI in general, and trust in a label in terms of a model's performance} (p.~152).
Figure \ref{fig:q4_reasons_for_trust} gives an overview over the different reasons or origins for trust, however interviewees usually did not specify whether they relate to labels or general AI systems.
Beyond the direct context of AI labels, interviewees also mentioned general AI skepticism (20 remarks) and regulation skepticism (8).
As an example, I1 reported very different views on AI in their company, \q{from euphorically enthusiastic to rather skeptically rejecting} (p.~30).

For trust in AI labeling, most comments were positive, for example I13 thinks that \q{it would help, yes. Because it's approved by professionals and trust is created} (p.~52).
However, I11, for example, doubts that \q{performance parameters help with such a question of trust} (p.~164), and I7 questioned whether he can truly \q{rely on such a label because that is such a specific thing} (p.~82), and said he would rather \q{test [models] himself} (p.~82).
In line with the established importance of institutions for increasing trust \cite{wischnewski2024seal} are the many comments regarding suitable labeling authorities.
Many interviewees appreciated the idea of having AI labels produced by a \q{central} (I8,~p.~124), \q{official} (I14,~p.~146), and \q{independent} (I15,~p.~122) authority, however, struggled to give a clear answer as to who could take this position.
Opposing this, nearly half of our interviewees raised concerns of subjectivity, as authorities could be \q{bribed} (I4,~p-~160) or possibly trick the labeling system for a more positive outcome, as it has happened with organic labels (I8,~p.~124).
Upon the question who should then certify AI, I9 responded fittingly: \q{quite democratically, the users} of labeling systems (p.~227).
This approach was greeted by mixed feelings, which can be seen from the remarks on self-certification versus third-party involvement -- I3 believes that \q{there must be something centralized, such that not everyone is allowed to make up their own label} (p.~184), yet others stated that having access to the certification framework \q{creates transparency and you can check [\dots] if it works as I imagine it will} (I5,~p.~252).
Placing performance at the core, I4 believed that \q{most people probably don't care [about properly understanding the system]. The main thing is that the end result is correct} (p.~180) - mirroring previous empirical results \cite{nussberger2022}.
I8 puts it similarly: \q{It somehow feels like what works well, what is well explained, counts more than who published it} (p.~116).

Exploring whether AI labels can be a means to create trust in AI necessitates to distinguish between different target audiences.
We found that our participants anticipated different trust requirements, depending on the trustees' levels of AI proficiency.
I11 saw the perspective of end-users' to be especially important, because \q{as an user of an AI, then of course I have the least trust} (I11,~p.~186). 
In that context, the label's effect was regarded as twofold:
On the one hand, concerns were raised that AI end-users might be overwhelmed or disinterested by the technicality of the metrics in the display (I15,~p.~106, I1,~p.~96).
On the other hand, it was positively remarked that metrics like power consumption made the AI model performance more understandable and tangible for users (I6,~p.~210).
Developers were remarked to inherently trust the systems they build: \q{I don't have a problem with trust in the sense that I'm the person who decides what kind of model to use} (I11,~p.~178).
Nonetheless, developers rely on use-case-specific explainability methods which they have to implement themselves in order to trust their model's outputs, as I8 reported. 

To answer our final research questions, trustworthiness is a broad problem with multiple dimensions and very personal views.
The two biggest factors for increasing trust seem to reside in responsible authorities and personal experience (i.e., from using available systems).
However, interviewees were not united in their positions as to who could possibly be a good authority and actively discussed the trustworthiness of authorities like companies, academia, open-source tools or governments.
The idea of receiving labels from an unbiased third-party authority was most popular, however open source access to labeling frameworks was also greeted.

\section{Discussion, Limitations, and Future Work}
The findings from this study reveal several key themes in the discussion of AI labeling practices, which relate back to our research questions.
In the following, we discuss these central points: the inherent trade-offs involved in designing labels (RQ2 \& RQ3), the potential of labels as nudges (RQ2), the ongoing challenge of trust in labels and their certifying authorities (RQ4), and lastly, the diverse needs and expectations of practitioners (RQ1 \& RQ3).
Each aspect presents both challenges and opportunities for improving AI transparency, communication, and trust.

A critical theme that emerged from our study revolves around the trade-off between simplicity and depth, as discussed in Section \ref{ssec:results_how_to_label}.
Generally, participants agreed on the need for simplicity, especially to facilitate quick decision-making and communication.
However, interviewees also expressed concerns of oversimplification and acknowledged the limitations of high-level labeling, especially when it comes to capturing the nuances of model performance and application suitability.
This reflects a broader tension in the field of AI communication: on the one hand, labels are meant to distill complex, often highly technical information into digestible, easily accessible formats; on the other hand, this simplification risks omitting essential details that could impact users' understanding and trust in the model’s capabilities.
This tension highlights a central challenge for how to design AI labeling systems: \textbf{labels must strike a balance between providing an overview that is both accessible and meaningful without sacrificing important detail.}
The desire for interactivity provides a potential solution to this dilemma.
The ability to adjust the importance of specific criteria based on user preferences could allow for a more dynamic, user-driven label experience.
This would enable users to engage with the label in a way that reflects their specific needs and for example prioritize predictive accuracy, resource efficiency, or interpretability.
In addition, it is important to acknowledge the connections between labeling and other forms of reporting.
By linking multiple representations, interested users can dive deeper into the intricacies which might reinforce their trust in labels.
By shaping reports and labels towards users' priorities, it may be possible to navigate the trade-off between simplicity and depth more effectively.

While not the focus of this study, it also became clear that another important role of AI labels is their potential as nudges, influencing user decisions by emphasizing certain aspects of model performance.
As our results in Section \ref{ssec:results_how_to_label} indicate, labels can function as a tool for guiding decision-making by drawing attention to key trade-offs between model attributes such as accuracy and energy consumption.
In this sense, \textbf{labels do more than simply present information -- they actively shape the decision-making process by highlighting the factors deemed most important.}

Trustworthiness was a recurrent topic in the interviews, with three central concerns as discussed in Section \ref{ssec:results_trustworthiness}: (1) trust in the label, (2) trust in the entity responsible for labels, and (3) label suitability for increasing trust.
Regarding the label's trustworthiness, participants highlighted the importance of clear, reliable metrics, but also expressed skepticism about the adequacy of labels to fully represent the complexity of AI models.
For experts, labels serve as a starting point for decision-making, but they are not a substitute for hands-on testing or exploring technical details.
The question of authoritative responsibility for labeling was contentiously discussed.
Participants suggested that a neutral, centralized authority (e.g., independent regulatory bodies or even academia) would lend legitimacy to AI labels, however concerns were raised regarding subjectivity and potential for bias.
Others advocated a more democratized approach, suggesting that developers themselves could play a key role in evaluating and certifying AI models. 
This underscores the difficulty of establishing trust in labeling systems, particularly as stakeholders may have competing interests in how AI systems are presented and evaluated. 
Following the growing trend of open source AI development and corresponding user-centric, community driven transparency could help in making labels trustworthy, however makes consistency and reliability all the more important.
For truly establishing them as a means to increase trust in AI, \textbf{labels need to be seen as part of a larger trust-building process that involves transparency, verifiability, and user experience.}

Lastly, the striking diversity in participants’ backgrounds, roles, and expertise as discussed in Section \ref{ssec:results_who_and_why} underscores the necessity for AI labels to be adaptable to different user groups and contexts.
Some develop their own AI models, however many others only interact with available AI services.
While labels are often seen as a promising tool for simplifying the communication of AI-related information, the broad spectrum of users, from technical experts to non-technical stakeholders, indicates that a "one-size-fits-all" approach would likely fall short.
Our results suggest that with any unified approach, \textbf{labels must allow for customization, ensuring that different audiences can extract the information they need.}
This need for adaptability aligns with previous research suggesting that AI reporting must consider varying audience expertise levels and roles \cite{fischer_towards_2024}.
In practice, this means that AI labels should incorporate flexibility, allowing users to choose the level of detail they wish to see.
In short, to become a useful tool, our study evidences that AI labels should balance accessibility and detail, shape decision-making by emphasizing key factors, support trust through transparency and verifiability, and enable customization for diverse audience needs.

While our study provides valuable insights, we also acknowledge potential limitations such as a sampling bias from recruiting participants via social media, which likely attracts those already interested in AI labeling while excluding skeptics. 
Additionally, creating and presenting the labels ourselves may have introduced response bias due to social desirability.
The visual similarities to energy and Nutri-score labels could have further influenced participants based on prior experiences with these systems.
Future research should validate findings with large-scale studies, test labels in real-world contexts, explore alternative designs and perspectives from skeptics, and focus on practical implementation and impact \cite{gansky_counterfacctual_2022}.

\section{Conclusion}

Our study highlights the multifaceted role of AI labeling in fostering trust and informed decision-making across different user groups.
We found evidence that AI labels are valuable due to their accessibility and potential to transfer knowledge, however must overcome challenges related to diverse audiences, technical comprehension, and metric transparency.
To maximize their impact, AI labeling systems should incorporate interactive features that allow for customization based on stakeholder priorities and knowledge.
Moreover, independent certification processes are essential to bolster trustworthiness.
By integrating these improvements, AI labels can serve as a cornerstone in the development of fair, accountable, and transparent AI systems, ultimately aligning technical advancements with societal expectations.

\bibliographystyle{unsrt}
\bibliography{references}

\newpage

\section*{Appendix}

\begin{figure}[h!]
    \centering
    \includegraphics[width=0.9\linewidth]{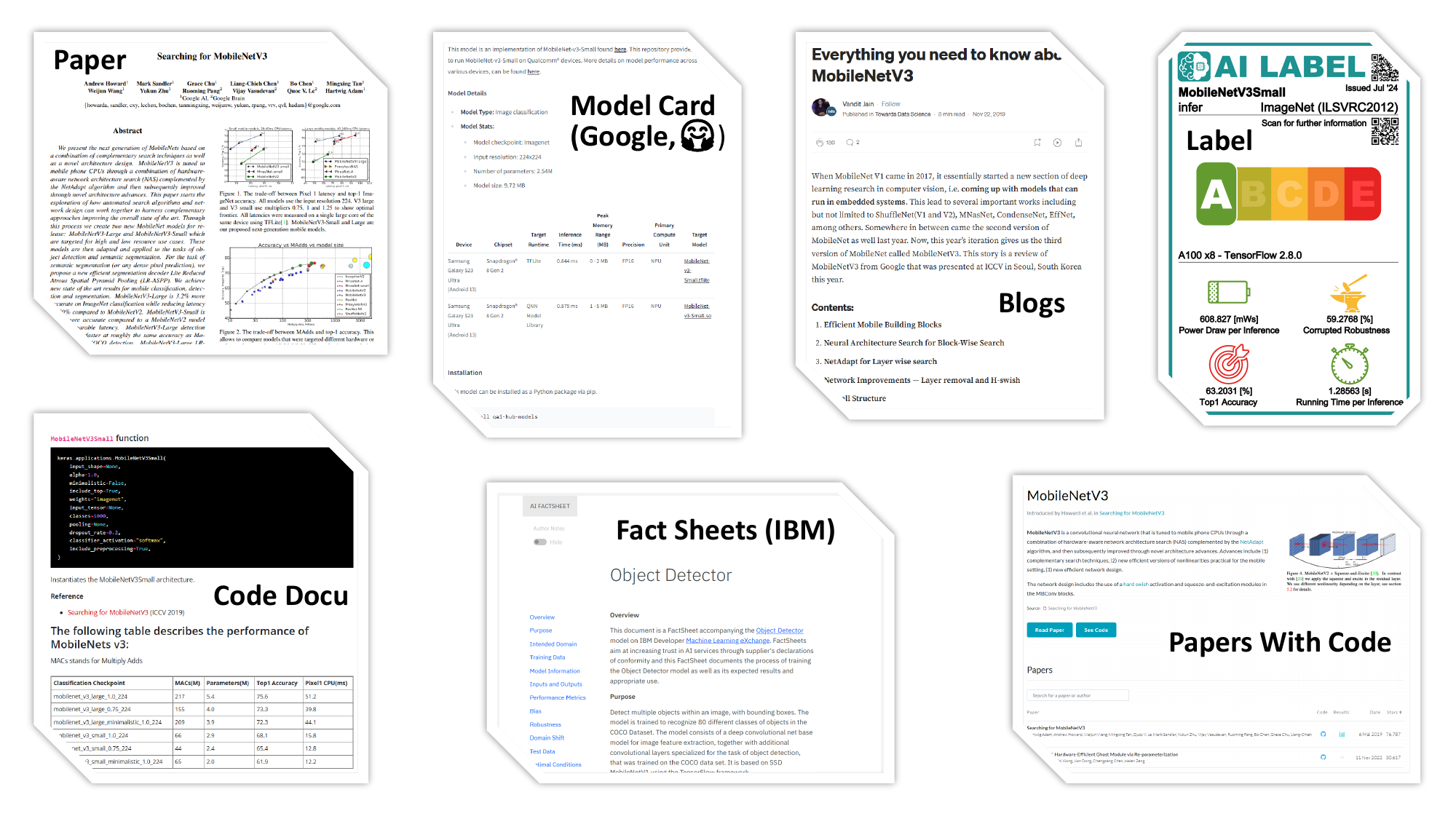}
    \caption{Different types of reporting on AI advances (here given for MobileNet \cite{Howard_2019_ICCV}), as discussed during interviews.}
    \label{fig:reporting_overview}
\end{figure}

\begin{figure}
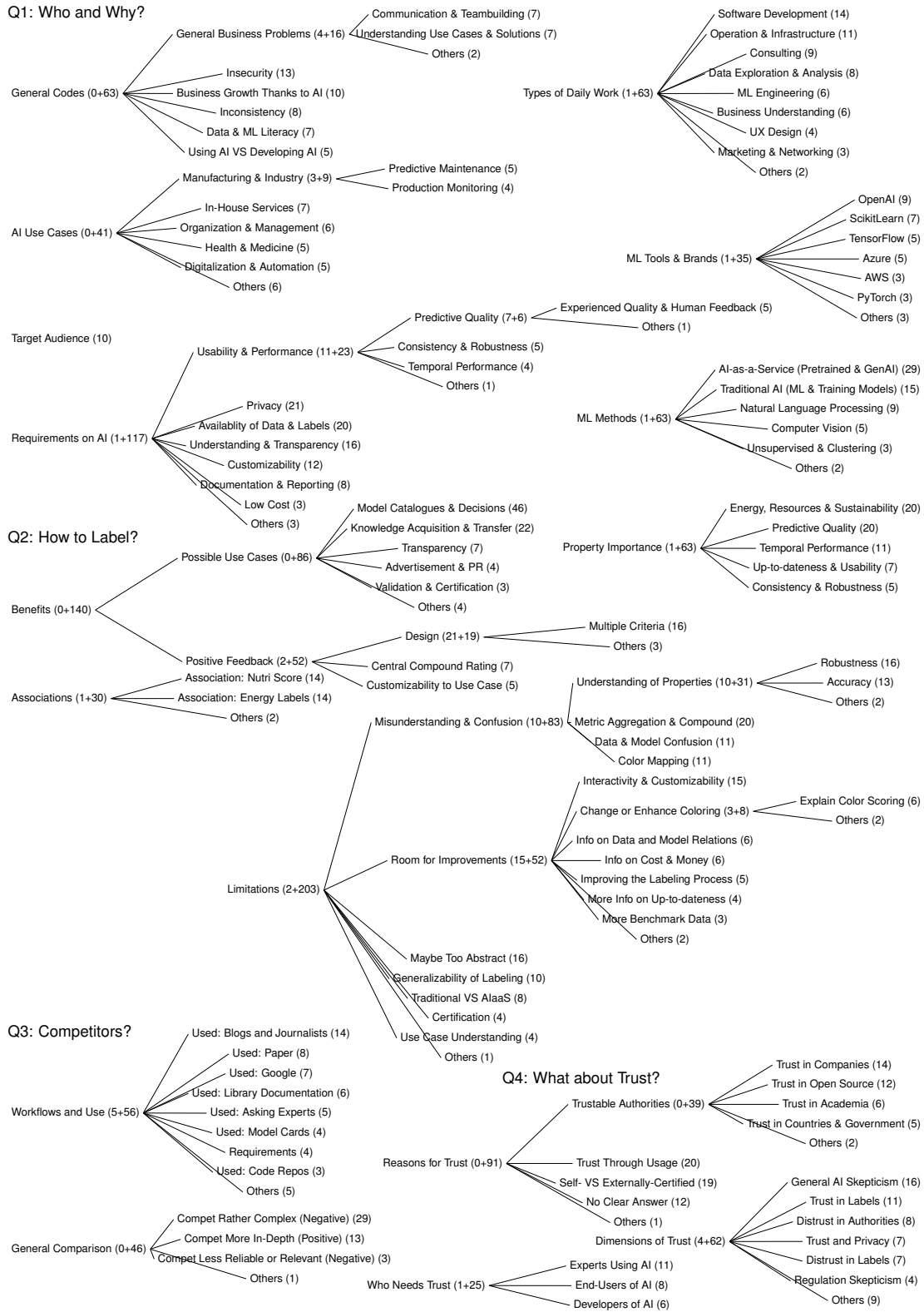

    \centering
    \resizebox{.9\textwidth}{!}{\begin{tikzpicture}[every node/.style={font=\sffamily, align=center},scale=1,transform shape]


        \node[anchor=north west] at (0, 0) {\Large Q1: Who and Why?};

        \node[anchor=north west] at (0, 0) {\begin{tikzpicture}[grow=right,level distance=180pt,scale=1,transform shape]
\Tree [.{General Codes (0+63)} [.{Using AI VS Developing AI (5)} ] [.{Data \& ML Literacy (7)} ] [.{Inconsistency (8)} ] [.{Business Growth Thanks to AI (10)} ] [.{Insecurity (13)} ] [.{General Business Problems (4+16)} [.{Others (2)} ] [.{Understanding Use Cases \& Solutions (7)} ] [.{Communication \& Teambuilding (7)} ] ] ]
\end{tikzpicture}};

        \node[anchor=north east] at (28, 0) {\input{material/codes_Q1_Types_of_Daily_Work}};

        \node[anchor=north west] at (0, -5) {\input{material/codes_Q1_AI_Use_Cases}};

        \node[anchor=north east] at (30, -11.5) {\input{material/codes_Q1_ML_Methods}};

        \node[anchor=north east] at (30, -6) {\input{material/codes_Q1_ML_Tools_and_Brands}};

        \node[anchor=north west] at (0, -9.5) {\begin{tikzpicture}[grow=right,level distance=180pt,scale=1,transform shape]
\Tree [.{Requirements on AI (1+117)} [.{Others (3)} ] [.{Low Cost (3)} ] [.{Documentation \& Reporting (8)} ] [.{Customizability (12)} ] [.{Understanding \& Transparency (16)} ] [.{Availablity of Data \& Labels (20)} ] [.{Privacy (21)} ] [.{Usability \& Performance (11+23)} [.{Others (1)} ] [.{Temporal Performance (4)} ] [.{Consistency \& Robustness (5)} ] [.{Predictive Quality (7+6)} [.{Others (1)} ] [.{Experienced Quality \& Human Feedback (5)} ] ] ] ]
\end{tikzpicture}};


        \node[anchor=north west] at (0, -17) {\Large Q2: How to Label?};
        
        \node[anchor=north west] (fig2) at (0, -16) {\begin{tikzpicture}[grow=right,level distance=180pt,scale=1,transform shape]
\Tree [.{Benefits (0+140)} [.{Positive Feedback (2+52)} [.{Customizability to Use Case (5)} ] [.{Central Compound Rating (7)} ] [.{Design (21+19)} [.{Others (3)} ] [.{Multiple Criteria (16)} ] ] ] [.{Possible Use Cases (0+86)} [.{Others (4)} ] [.{Validation \& Certification (3)} ] [.{Advertisement \& PR (4)} ] [.{Transparency (7)} ] [.{Knowledge Acquisition \& Transfer (22)} ] [.{Model Catalogues \& Decisions (46)} ] ] ]
\end{tikzpicture}};

        \node[ anchor=north east] (fig3) at (30, -21) {\begin{tikzpicture}[grow=right,level distance=180pt,scale=1,transform shape]
\Tree [.{Limitations (2+203)} [.{Others (1)} ] [.{Use Case Understanding (4)} ] [.{Certification (4)} ] [.{Traditional VS AIaaS (8)} ] [.{Generalizability of Labeling (10)} ] [.{Maybe Too Abstract (16)} ] [.{Room for Improvements (15+52)} [.{Others (2)} ] [.{More Benchmark Data (3)} ] [.{More Info on Up-to-dateness (4)} ] [.{Improving the Labeling Process (5)} ] [.{Info on Cost \& Money (6)} ] [.{Info on Data and Model Relations (6)} ] [.{Change or Enhance Coloring (3+8)} [.{Others (2)} ] [.{Explain Color Scoring (6)} ] ] [.{Interactivity \& Customizability (15)} ] ] [.{Misunderstanding \& Confusion (10+83)} [.{Color Mapping (11)} ] [.{Data \& Model Confusion (11)} ] [.{Metric Aggregation \& Compound (20)} ] [.{Understanding of Properties (10+31)} [.{Others (2)} ] [.{Accuracy (13)} ] [.{Robustness (16)} ] ] ] ]
\end{tikzpicture}};

        \node[ anchor=north east] (fig3) at (30, -16) {\input{material/codes_Q2_Property_Importance}};

        \node[ anchor=north west] (fig3) at (0, -21.5) {\input{material/codes_Q2_Associations}};

        \node[ anchor=north west] (fig3) at (0, -10.5) {\input{material/codes_Q2_Target_Audience}};


        \node[anchor=north west] at (0, -33) {\Large Q3: Competitors?};
        
        \node[ anchor=north west] (fig3) at (0, -33) {\input{material/codes_Q3_Workflows_and_Use}};

        \node[ anchor=north west] (fig3) at (0, -39) {\input{material/codes_Q3_General_Comparison}};


        \node[anchor=north west] at (16, -34.5) {\Large Q4: What about Trust?};

        \node[ anchor=north east] (fig3) at (30, -34) {\begin{tikzpicture}[grow=right,level distance=180pt,scale=1,transform shape]
\Tree [.{Reasons for Trust (0+91)} [.{Others (1)} ] [.{No Clear Answer (12)} ] [.{Self- VS Externally-Certified (19)} ] [.{Trust Through Usage (20)} ] [.{Trustable Authorities (0+39)} [.{Others (2)} ] [.{Trust in Countries \& Government (5)} ] [.{Trust in Academia (6)} ] [.{Trust in Open Source (12)} ] [.{Trust in Companies (14)} ] ] ]
\end{tikzpicture}};

        \node[ anchor=north east] (fig3) at (22, -40.5) {\input{material/codes_Q4_Who_Needs_Trust}};

        \node[ anchor=north east] (fig3) at (30, -37.8) {\input{material/codes_Q4_Dimensions_of_Trust}};
        
    \end{tikzpicture}
}
    \caption{Full display of our code system. Numbers indicate how often the code was used (+ the number of used subcodes).}
    \label{fig:full_code_system}
\end{figure}

As additional material for our work, we here feature two figures (note that more supplementary material can be found in our repository at \url{www.github.com/raphischer/labeling-evaluation}).
Firstly, Figure \ref{fig:reporting_overview} showcases the established forms of AI reporting in contrast to labeling.
This comparison was shown and discussed in the third part of our interviewees to answer research question RQ3.
Figure \ref{fig:full_code_system} depicts our complete code system in all its complexity, subdivided into the four central research directions.
The codes are also annotated with the number of occurrences (+ number of occurring subcodes), from which the first level codes were already displayed in Table \ref{tab:codesystem_overview}.

\end{document}